\documentclass[letterpaper, 10 pt, conference]{ieeeconf}  

\IEEEoverridecommandlockouts                              

\usepackage{graphics} 
\usepackage{epsfig} 
\usepackage{mathptmx} 
\usepackage{times} 
\usepackage{amsmath} 
\usepackage{amssymb}  
\usepackage{balance}

\title{\LARGE \bf A Neurorobotic Experiment for Crossmodal \\ Conflict Resolution in Complex Environments *}

\author{German I. Parisi$^1$, Pablo Barros$^1$, Di Fu$^{1,2,3}$, Sven Magg$^{1}$, Haiyan Wu$^{2,3,4}$, Xun Liu$^{2,3}$, Stefan Wermter$^1$
\thanks{$^{1}$Knowledge Technology, Department of Informatics, Universit\"at Hamburg, Germany.
        {\tt\small \{surname\}@informatik.uni-hamburg.de}}%
\thanks{$^{2}$CAS Key Laboratory of Behavioral Science, Chinese Academy of Sciences (CAS), Beijing, China. {\tt\small \{fud,wuhy,liux\}@psych.ac.cn}}%
\thanks{$^{3}$Department of Psychology, University of CAS, Beijing, China.}%
\thanks{$^{4}$Division of the Humanities and Social Sciences, Caltech, CA, USA.}
\thanks{*Open-source code: {\tt\small cml.knowledge-technology.info} }
}

\begin{document}

\maketitle
\thispagestyle{empty}
\pagestyle{empty}

\begin{abstract}
Crossmodal conflict resolution is crucial for robot sensorimotor coupling through the interaction with the environment, yielding swift and robust behaviour also in noisy conditions.
In this paper, we propose a neurorobotic experiment in which an iCub robot exhibits human-like responses in a complex crossmodal environment.
To better understand how humans deal with multisensory conflicts, we conducted a behavioural study exposing 33 subjects to congruent and incongruent dynamic audio-visual cues.
In contrast to previous studies using simplified stimuli, we designed a scenario with four animated avatars and observed that the magnitude and extension of the visual bias are related to the semantics embedded in the scene, i.e., visual cues that are congruent with environmental statistics (moving lips and vocalization) induce the strongest bias.
We implement a deep learning model that processes stereophonic sound, facial features, and body motion to trigger a discrete behavioural response.
After training the model, we exposed the iCub to the same experimental conditions as the human subjects, showing that the robot can replicate similar responses in real time.
Our interdisciplinary work provides important insights into how crossmodal conflict resolution can be modelled in robots and introduces future research directions for the efficient combination of sensory observations with internally generated knowledge and expectations.
\end{abstract}
\section{INTRODUCTION}

Robots operating in the real world must efficiently interact with their surroundings.
Similarly to biological agents, robots can make use of an array of sensors for processing multiple modalities such as vision, audio, haptics, and proprioception with the goal of promptly undertaking behaviourally relevant decisions through the combination of sensory observations with prior knowledge and expectations (e.g. internally generated models of the world)~\cite{Noda14}.

The integration of multisensory information has been widely studied in the literature, e.g. in terms of the learning of multisensory representations from heterogeneous sensor data~\cite{Vavrecka2013} or through modelling the development of sensorimotor skills in embodied autonomous agents~\cite{Hwang17}.
However, it is typically assumed that the multisensory measurements provide a complete and coherent data stream that can be straightforwardly integrated on the basis of spatial and/or temporal coincidence.
Nevertheless, behaviour should be swift and singular also in situations of sensory uncertainty and multisensory conflict~\cite{Polley2017}.

Multisensory conflicts result from sensory uncertainty and variable reliability under noisy environmental conditions~\cite{CruzIROS}.
For instance, when having to identify a speaker in a room, auditory information may become unreliable if there is music playing in the background, and localizing the source of sound may be difficult in this condition.
However, humans have learned from environmental statistics that seeing moving lips is typically associated with a person speaking.
Therefore, this expectation will influence our ability to promptly detect a speaker in a noisy room.
Conversely, audio may be more reliable for detecting a speaker if the light conditions of the room are adverse for vision.

The mammalian brain comprises multisensory areas that integrate information on the basis of low-level stimuli properties (e.g., spatial and temporal coincidence) and high-level properties such as spatiotemporal congruency, prior knowledge, and expectations~\cite{Malacuso2005}.
The interplay of low-level (bottom-up) and high-level (top-down) information is crucial for swift decision making and conflict resolution~\cite{Polley2017}.
A number of behavioural studies have shown different audio-visual effects that reflect the intricate interplay of bottom-up and top-down information processing.
Widely studied illusions are the spatial ventriloquism effect~\cite{Jack73}, where the auditory stimulus is perceptually shifted towards the position of the synchronous visual one, and the McGurk effect~\cite{mcgurk76}, in which a mismatched audio-visual stimulus of vocalizing a sound leads to the perception of an illusory sound.

The aforementioned studies with human subjects have provided valuable insights into how multisensory stimuli can be modelled in artificial systems.
However, the stimuli used for triggering responses do not reflect the complexity of the environment that artificial agents are expected to interact with.
Critically, audio-visual spatial tasks typically use (over)simplified stimuli such as light blobs and sound clicks, and show only one stimulus per modality~(e.g.,~\cite{Kording17}).
Under these experimental conditions, subjects produce responses based on the spatiotemporal congruency of the audio-visual cues but neglect likewise important factors such as semantic congruency and expectations.
These top-down factors significantly contribute to the development of a robust percept~\cite{Zhu2017} and are crucial for modelling multisensory integration and conflict resolution in robots.

In this paper, we present a neurorobotic study of crossmodal conflict resolution in a complex environment.
In order to better understand how humans solve crossmodal conflicts, we extend a previously proposed behavioural study consisting of an audio-visual spatial localization task~\cite{ParisiICDL}.
Our novel study was conducted in an immersive projection environment and comprises a scene with four animated avatars sitting around a table which can produce congruent and incongruent audio-visual stimuli.
We trained a deep learning model to trigger human-like responses and evaluated this approach with an iCub robot exposed to the same experimental conditions as human subjects.

Our neurorobotic study contributes to the leverage of current models of robot perception and behaviour taking into account the complex nature of crossmodal environments and the way humans perceive, learn, and act on the basis of rich (and often uncertain) streams of multisensory input.
The main contribution of this work is twofold.
First, we provide a quantitative analysis of visually-induced bias on the estimation of sound source localization for different types of audio-visual conflicts.
Our findings suggest that i) semantics embedded in the scene modulate the magnitude and extension (in terms of integration window) of the visually-induced bias and ii) expectation-driven perceptual mechanisms introduced by the exposure to animated avatars induces a systematic error in the responses comprising static avatars.
Second, we implement a deep neural network architecture that models human-like behaviour and is shown to trigger similar responses with an iCub robot in real time.
The model is motivated by neuroscientific findings suggesting i) the processing of auditory cues (sound source localization) and visual cues (face and body motion) in distinct brain areas and ii) their combination, in terms of neurons responding to (in)congruent multisensory representations, in higher-level areas~\cite{Dahl2010}.

\section{BEHAVIOURAL STUDY}

\subsection{Overview}

In a previous study~\cite{ParisiICDL}, we proposed an audio-visual (AV) spatial localization task that comprised a set of 4 animated avatars.
The AV stimuli consisted of one avatar with moving lips along with a synchronous, spatially congruent or incongruent auditory cue.
Our findings suggest that human subjects were more inaccurate to spatially localize the sound when exposed to incongruent AV stimuli.
This study made a step towards evaluating multisensory conflict resolution in complex environments, i.e., going beyond the typically used simplified AV stimuli such as lights and clicks~\cite{Kording17}.
However, it is subject to a number of limitations.
First, we tested subjects on AV stimuli comprising one visual and one auditory cue.
Crucially, natural scenes may include multiple visual cues influencing multisensory integration to different extents.
Therefore, it is important to assess the interplay of multiple visual cues conveying different semantic meaning, e.g., lip and body movement.
Second, the visual stimuli were displayed on a 17-inch monitor and the auditory ones were presented via a headphone set, thus significantly differing from natural crossmodal environments and the way humans (and robots) interact with their surroundings.

In this novel study, we extend our experimental design to test new hypotheses and propose a new immersive experimental setup so that human subjects and an iCub humanoid robot can be exposed to the same experimental conditions.

\begin{figure*}[t]
\centering
\includegraphics[width=\textwidth]{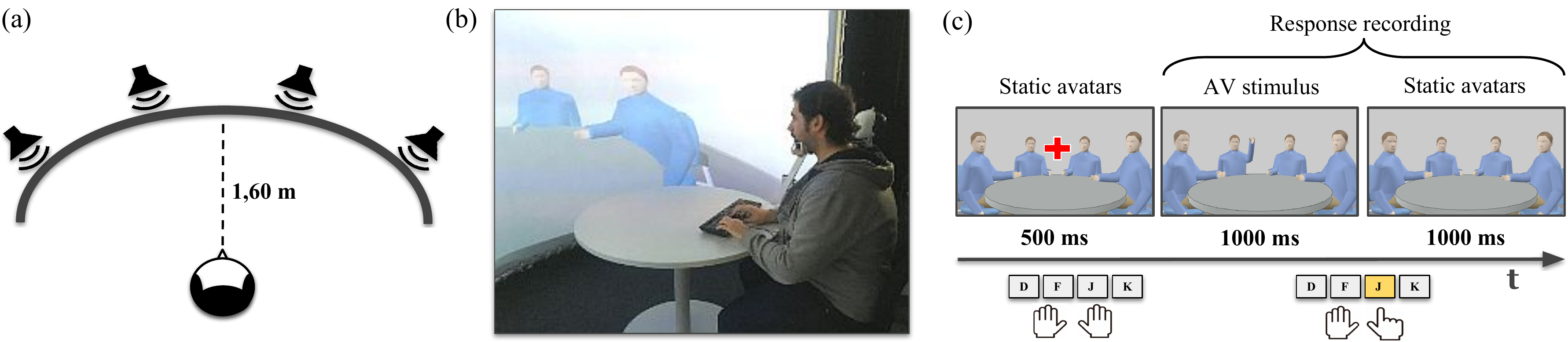}
\caption{Behavioural study on audio-visual localization: (a) Immersive experimental setup with acoustically transparent concave projection screen (b) Subject selecting a position via a keyboard (c) Schematic illustration of one trial of the AV localization task.}
\label{fig:experiment}
\end{figure*}


\subsection{Participants}

A total of 33 subjects (7 female, aged 21--32, right-handed) participated in our experiment.
All participants reported that they did not have a history of any neurological conditions (seizures, epilepsy, stroke), and had normal or corrected-to-normal vision and hearing.
This study was conducted in accordance to the principles expressed in the Declaration of Helsinki.
Each subject signed a consent form approved by the Ethics Committee of Universit\"at Hamburg.
The range of accuracy of the task was $28\%-98\%$ (mean $64\% \pm 23\%$).

\subsection{Apparatus, Stimuli, and Procedure}

Participants sat at a desk with their chins on a chin-rest positioned 160cm from the projection screen~(see Fig~\ref{fig:experiment}.a). 
Visual stimuli were presented on a concave projection screen made of acoustically transparent cloth encompassing much of the visual field.
The semi-circular screen has a diameter of 2.42m and a height of 2.72m of which 2.38m were illuminated.
The visual stimuli were presented overhead from 4 ceiling-mounted HD-projectors with individual projections warping-corrected and blended to produce a single continuous image~(Fig~\ref{fig:experiment}.b).
The mouths of the avatars were 1.36m above the ground and the avatar's locations were at -33, -11, +11, and +33 degrees off the centre (fixation point).
Behind the screen were 4 free-field speakers that differed in azimuth and were placed to match the mouths of the projected avatars.
Due to the complexity of the environment which creates stimulus onset delays for both visual and auditory streams, the onsets have been synced using a high-speed camera (1000 fps).
After synchronization, the average audio-visual onset difference is 0 $\pm$ 32ms.

The AV localization task consisted of the subjects having to select which avatar (out of the 4 avatars in the scene) they believe the auditory cue is coming from.
The 4 avatars may move their lips and/or arm in temporal correspondence with an auditory cue.
The latter consists of a vocalized combination of 3 syllables (all permutations without repetition composed of "ha", "wa", "ba").
The duration of both visual and auditory stimuli is 1000 ms.

The experiment comprised 5 AV conditions:
\begin{enumerate}
\item \textbf{Baseline}: Auditory cue and static avatars.
\item \textbf{Moving Lips}: Auditory cue and one avatar with moving lips.
\item \textbf{Moving Arm}: Auditory cue and one avatar with a moving arm.
\item \textbf{Moving Lips+Arm}: Auditory cue and one avatar with moving lips and arm.
\item \textbf{Moving Lips--Arm}: Auditory cue and one avatar with moving lips and another avatar with a moving arm.
\end{enumerate}

For all the conditions except for \textbf{Condition 1}, the AV pair may be spatially congruent or incongruent.
In \textbf{Condition 5}, spatial congruency comprises lips-audio or arm-audio pairs.
If we consider all the AV-pair combinations (congruent and incongruent) derived from the 5 conditions, it results in 200 trials.

Participants began the experiment with 12 practice trials composed of congruent AV stimuli.
This practice session ensured that participants understood the instructions and were using the keyboard properly to select one of the 4 locations.
The formal task consisted of 600 trials (3 sessions of 200 trials presented in random order).
A schematic illustration of one trial is shown in~Fig~\ref{fig:experiment}.c.
Each trial started with static avatars and a fixation point for 500 ms, followed by an AV stimulus and then another 1000 ms with static avatars.
The subjects were asked to produce a response within 2000 ms after the onset of the AV stimulus.
After completing the experiment, the subjects were asked which perceptual strategy they believe they adopted to solve the AV task: i) mostly auditory, ii) mostly visual, or iii) mixed.

\subsection{Results and Analysis}

We analyzed the obtained behavioural data in terms of the error rate (ER) with respect to the ground-truth position of the auditory cue.

The amount of visually-induced bias on auditory cues depends on the proximity of the cues and their position with respect to the field of view.
In the spatial ventriloquism effect, the perception of the auditory stimulus is shifted towards the direction of the visual cue in relation to their spatial proximity.
This integration window, however, breaks down when the distance between the two stimuli is greater than 20-25 degrees and the magnitude of the visual bias becomes negligible~\cite{Magosso12}.
Furthermore, visual spatial resolution is higher in the center of the field of view (FOV), thus the magnitude of the visual bias is expected to be higher towards the center rather than towards the periphery~\cite{Odegaard15}.
Consequently, we analyzed the ER by taking into account the distance between the avatars and their absolute location with respect to the FOV of the observer.
For this purpose, we divided the conditions in terms of (mirrored) spatial relationships:
\begin{itemize}
\item \textbf{Congruent}: Congruent AV pair from one avatar.
\item \textbf{Central}: Incongruent AV pair from the two avatars in the center.
\item \textbf{Lateral}: Incongruent AV pair from two of the avatars on the right or the left side.
\item \textbf{1-Avatar Gap}: Incongruent AV pair from avatars having a 1-avatar gap.
\item \textbf{2-Avatar Gap}: Incongruent AV pair from avatars having a 2-avatar gap, i.e., the two at each side of the screen.
\end{itemize}

\begin{figure*}[t]
\centering
\includegraphics[width=0.85\textwidth]{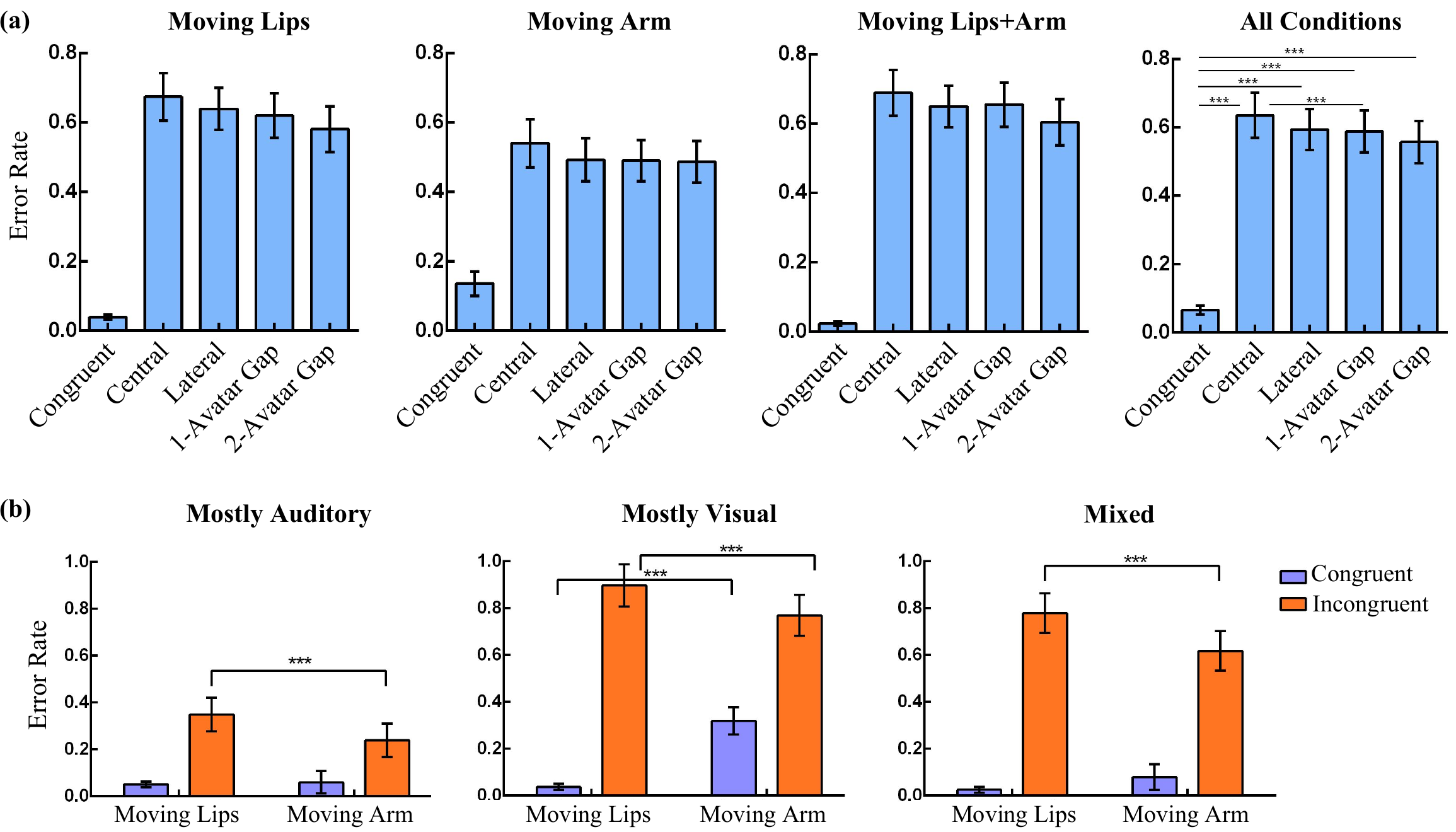}
\caption{Data analysis of (in)congruent AV stimuli in terms of the error rate (ER) with respect to the ground-truth position of the auditory cue. (a) Analysis using the distance between the avatars and their position within the FOV of the observer.  (b) ER according to the different perceptual strategies reported by the subjects. (* denotes $0.01<p<0.05$, ** $0.001<p<0.01$, and *** $p<0.001$). }
\label{fig:experiment_results}
\end{figure*}

We conducted a two-way repeated measures analysis of variance (ANOVAs).
The ER for the different conditions is shown in Fig.~\ref{fig:experiment_results}.a.
Across the conditions, the effect of the distance between the avatars was significant ($F=37.08$, $p<0.001$, $\eta^2=~0.70$).
Post-hoc \textit{t}-tests revealed that the congruent condition had the lowest error rate (ER=7\%), while the highest error was for the avatars in the center (ER$=~64\%$, $p<0.01$) followed by a 1-avatar gap (ER$=59\%$, $p<0.01$), lateral (ER$=59\%$, $p<0.14$), and a 2-avatar gap (ER$=56\%$, $p<0.05$).
Furthermore, we analyzed the ER across the 3 sessions.
We found a significant learning effect (LE) between sessions 1 and 2 ($p<0.05$), while the learning effect between sessions 2 and 3 was not significant ($p<0.71$).
The interaction between LE and avatar distance was not significant ($F=1.36$, $p=0.24$, $\eta^2=0.04$).

We analyzed the responses in terms of the adopted perceptual strategy adopted by the subjects (Fig.~\ref{fig:experiment_results}.b).
After the experiment, 14 subjects reported to have used a mostly auditory strategy (AS), 9 subjects a mostly visual strategy (VS), and 10 subjects a mixed strategy~(MS).
Subjects using an AS had the highest accuracy (ER$=20\%$) with respect to VS (ER$=54\%$, $p<0.001$) and MS (ER$=43\%$, $p<0.01$).
The main effects of the strategy ($F=14.65$, $p<0.001$, $\eta^2=0.49$), the visual cue ($F=48.00$, $p<0.001$, $\eta^2=0.62$), and the congruency of the stimulus ($F=192.06$, $p<0.001$, $\eta^2=0.87$) were significant.
The interaction among these three factors was also significant ($F=6.06$, $p< 0.001$, $\eta^2=0.29$).

Our findings suggest that the embedded semantics significantly modulate the magnitude and extension (in terms of integration windows) of the visually-induced bias.
Moving lips cause higher error rates in the final estimate of the location of the sound with respect to a moving arm (which is visually more salient).
This is in line with fMRI studies suggesting the highest multisensory integration in terms of neural activation for congruent mouth-voice stimuli~(e.g.,\cite{Zhu2017}).
In contrast to previous studies showing that the integration window breaks down for distances greater than 20-25 degrees~\cite{Magosso12}, in our case the magnitude of the visual bias is significant also when the incongruent AV stimuli are coming from the two avatars at the extremes of the screen.
This can be interpreted in terms of synchronized AV pairs being merged as a single event irrespective of their spatial disparity due to their temporal correlation perceived as a form of causation~\cite{Parise2012}.
We can argue that embedded semantics in the scene contribute to a wider integration window with respect to the boundaries empirically found for simplified stimuli~\cite{Dahl2010}.
To further verify this hypothesis, we examined whether the induced bias during incongruent AV stimuli was in the direction of the visual cue, i.e., we tested the ventriloquist effect.
The results for the different conditions are shown in Fig.~\ref{fig:experiment_results2}.d, which are consistent with the hypothesis that visual cues encoding environmental statistics induce a stronger bias and that the magnitude of the bias is related to the embedded semantics, e.g., \textit{moving lips+arm} induces a slightly stronger bias than \textit{moving lips}.
Furthermore, while an incongruent arm movement in \textit{moving lips--arm} acts as distractor decreasing the magnitude of the visual bias towards the lips, this magnitude is still significant.

Finally, when the auditory cue is played along with static avatars, subjects were not as accurate as expected in the absence of a visual bias.
%
One hypothesis for this effect is that the extended exposure to animated avatars may create the expectation of seeing similar animated patterns in the next trials, thus perceiving a static avatar as incongruent with respect to an expected dynamic visual cue.
However, a more extensive study is required to verify this hypothesis and measure the modulatory effects of expectation learning~\cite{Diehl2014}.

\section{NEUROROBOTIC EXPERIMENT}

The goal of the neurorobotic experiment was to trigger human-like responses with an iCub~\cite{Metta2010} exposed to the same conditions as the human subjects.
For this purpose, we used the collected behavioural data to train a deep learning model and compared the results with human responses.

\subsection{Multichannel Deep Learning Model}

The modelling of crossmodal integration and conflict resolution is particularly complex since behaviourally relevant responses are mediated by the interplay of functionally distinct brain areas~\cite{Dahl2010}.
For instance, the superior colliculus in the midbrain processes stimuli on the basis of their spatiotemporal alignment, whereas higher-order areas, such as the superior temporal sulcus in the temporal cortex, process feature-based representations and activate strongly for semantically congruent stimuli.

\begin{figure}[t]
\centering
\includegraphics[width=0.47\textwidth]{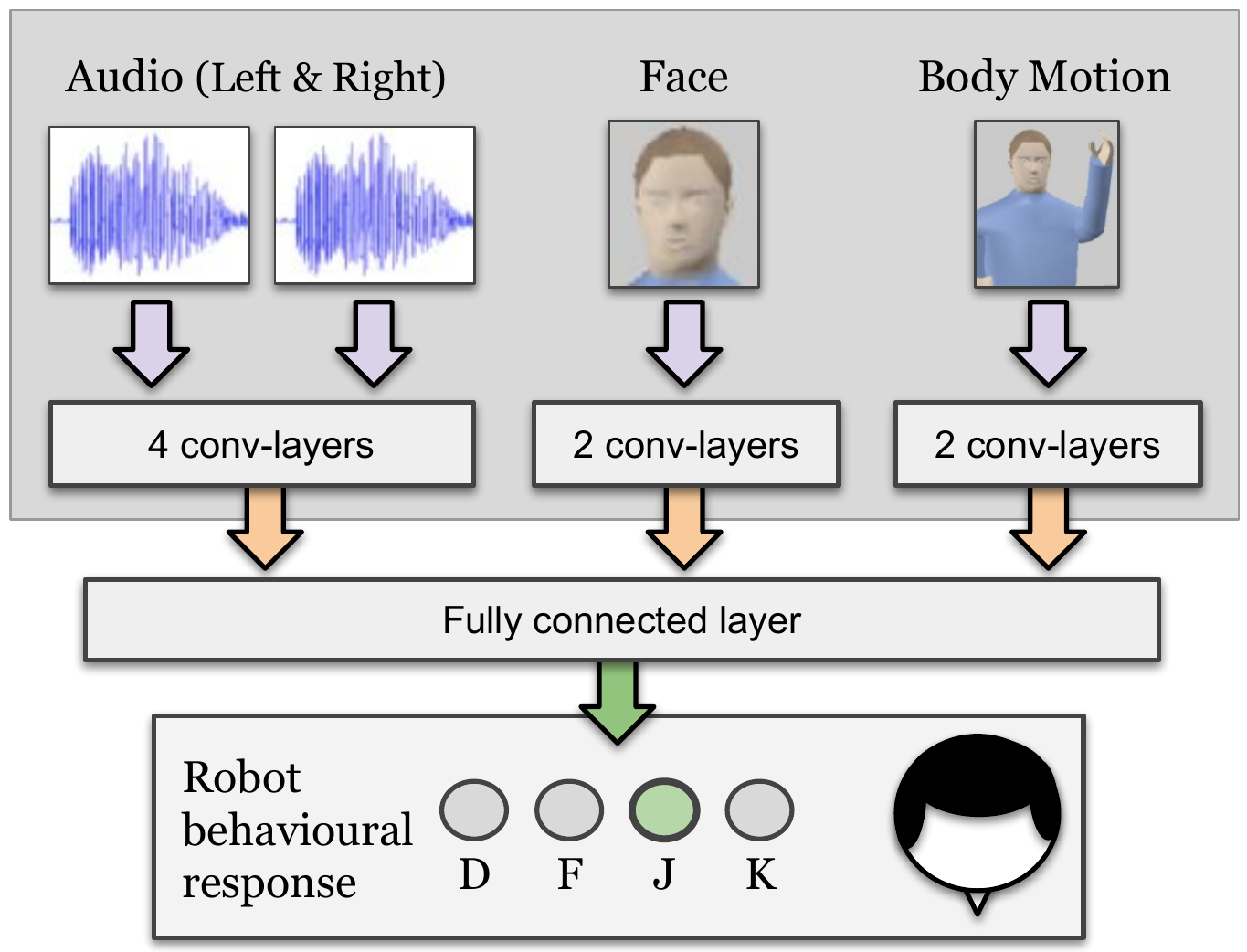}
\caption{Multichannel deep learning model for multisensory integration and conflict resolution. The model combines sound localization, facial features, and body motion to produce a discrete behavioural response in real time. Each channel is first trained with modality-specific spatial information (gray bounding box) and used as input for a hidden layer trained with multisensory representations using human behavioural responses as the teaching signals.}
\label{fig:multis}
\end{figure}

We propose a deep learning model processing both spatial and feature-based information in which low-level areas (such as the visual and auditory cortices) are predominantly unisensory, while neurons in higher-order areas encode multisensory representations.
The proposed architecture comprises 3 input channels (audio, face, and body motion) and a hidden layer that computes a discrete behavioural response on the basis of the output of these unisensory channels (see Fig.~\ref{fig:multis}).



For the audio, we extract each auditory channel from the file recorded with the binaural microphone of the robot and apply the short-time Fourier transform, obtaining one spectrogram for each auditory channel with a dimension of 512x26, i.e., 512 temporal bins with 26 descriptors each.
For the body movement channel, we average the images from the RGB camera over the whole stimuli duration and create one 80x60 grayscale image representation with the body movement of all the avatars.
Similarly, for the face channel we average each face individually and obtain four averaged images per stimulus, each of them transformed to grayscale and resized to 120x120 pixels.

Our network is composed of three convolutional channels, following a multichannel architecture~\cite{Barros2016}: one for learning auditory features, a second for body movement and the third for facial expressions.
The auditory channel is composed of four pairs of convolution layers, each with 3x3 filters and the last layer of each pair has a 2x2 stride.
The first pair of layers has 8 filters, the second 16, the third 24 and the last one 32.
The face and body motion channels comprise two convolution layers, each with 16 filters of size 3x3 followed by a max pooling layer with a receptive field of 2x2 and a dropout layer. 
The same architecture is used by our face channel.
These hyperparameters were selected through an exhaustive exploratory search for optimizing the source location of unisensory stimuli.

To train our model on crossmodal conflict resolution, we first train the individual channels using modality-specific spatial information (Fig.~\ref{fig:multis}; gray bounding box).
The auditory channel is trained to locate a sound source, the face channel to locate moving lips, and the body motion channel to locate arm movement.
This procedure ensures that each channel is able to describe modality-specific stimuli.
After the individual training of these 3 channels, a fully connected hidden layer receives modality-specific representations as input and is trained using the human responses as teaching signals.
The output softmax layer represents a probability distribution over the 4 possible responses.

\subsection{Robot Behaviour}

\begin{figure}[t]
\centering
\includegraphics[width=0.48\textwidth]{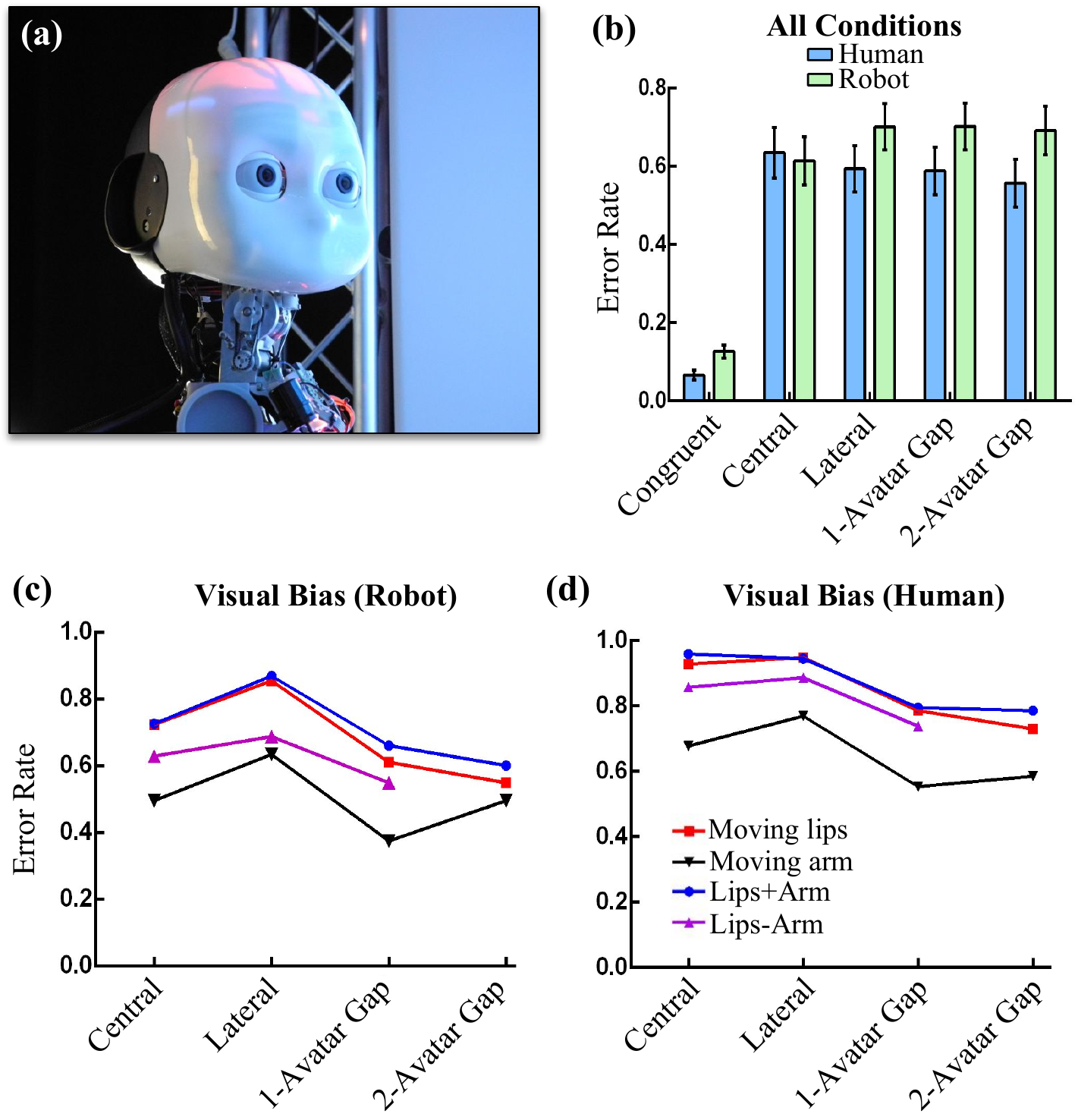}
\caption{Neurorobotic experiment: (a) iCub exposed to the immersive experimental setup (b) Error rate after exposure to congruent and incongruent AV stimuli (see Fig.~\ref{fig:experiment_results}.a for comparison with human subjects) (c) Robot vs Human comparison in terms of the ventriloquism effect (response  biased towards the visual cue) showing similar trends for all the conditions with incongruent AV stimuli.}
\label{fig:experiment_results2}
\end{figure}

For a direct human-robot comparison, we placed the iCub in front of the projection screen~(Fig.~\ref{fig:experiment_results2}.a; see Fig.~\ref{fig:experiment}.a for setup with humans).
In order to prevent biasing the robot behaviour towards a specific subject, we evaluate the model using \textit{leave-one-out} cross-validation with the responses of 32 subjects for training and of 1 subject for testing.
Since each participant produced 600 responses, we had 32$\times$600 training data points for each training fold, for which a new network was initialized.
This training procedure resulted in 33 network instances from which we generated 33$\times$600 responses used to compare robot-vs-human behaviour.

The error rates of the robot averaged across all conditions are shown in Fig.~\ref{fig:experiment_results2}.b~(see Fig.~\ref{fig:experiment_results}.a for comparison with humans), where it can be seen that the human-vs-robot ER difference is not significant ($F=1.303,p=0.26,\eta^2=0.02$).
Interestingly, there is an inverse trend with respect to humans in which the ER is higher for the avatars in the center and decreases for the ones at the sides.
This difference can be explained due to the different ways in which humans and the robot process incoming visual input.
Human vision has higher spatial resolution towards the center of the FOV (referred to as foveal vision), which leads to a stronger visual bias over the estimate of the sound source's location when the visual cue occurs towards the center~\cite{Odegaard15}.
This effect in relation to the position of the visual cue is shown in Fig.~\ref{fig:experiment_results2}.b, where the ER is higher for the \textit{central} position with a decreasing trend for \textit{lateral}, \textit{1-avatar gap}, and \textit{2-avatar gap}.
On the other hand, the visual input processed by the robot does not comprise such foveal property, and consequently, the visual bias has the same magnitude irrespective of its position within the FOV of the camera.
However, since the model is trained with data collected from robot sensors but using human responses as a teaching signal, there is a compensation artefact introduced by the hidden layer which results in such an inverse trend of the magnitude of the visual bias in relation to its position.
In order to address this artefact, it would be necessary to model properties of foveal vision embbeded in the convolutional channels processing the visual input~(e.g.,~\cite{Luo17}).

In terms of the magnitude of the bias reflecting environmental statistics, we analyzed the proportion of ER due to shifting the estimate towards the visual cue (ventriloquism effect).
It can be seen from Fig.~\ref{fig:experiment_results2}.c-d that the behaviour of the robot resembles human responses for all the conditions.

\section{FUTURE WORK}

The obtained results motivate further research in three main directions.
First, the experimental scenario can be extended to more natural scenes, e.g., by displaying real-world videos with human characters.
Here, we kept the appearance of the animated avatars equal to prevent any kind of association or identity learning effect.
However, human characters or human-like avatars can have significantly more degrees of freedom in terms of body movement, expressiveness, and identity features.
In this case, several factors would be more difficult to control in a systematic way.

Second, the deep learning model was trained in a supervised fashion, i.e., by providing the expected responses as a target.
Instead, it may be of interest to study whether and how such behaviour can emerge from the unsupervised exposure to congruent AV stimuli, e.g., by learning environmental statistics.
This research direction may be of particular interest for neurorobotic approaches aimed to model specific areas of the brain responsible for multisensory interaction and conflict resolution.

Third, we observed that subjects adopting a strategy that relied mostly on auditory cues exhibited smaller error rates.
Studies suggest that the brain changes its strategy according to the reliability of sensory drive and mechanisms of cognitive control~\cite{Yang2017}.
Consequently, it would be of interest to further study and model the dynamic selection of perceptual strategies on the basis of modality-specific reliability, conflict adaptation effects, top-down attention, and prior knowledge.


\section{CONCLUSIONS}

We introduced and evaluated a deep learning architecture for modelling human-like responses with an iCub robot exposed to complex audio-visual stimuli.
For the modelling of behavioural responses that functionally reflect crossmodal conflict resolution in humans, we designed an AV localization task comprising a set of congruent and incongruent conditions with four animated avatars.
This scenario provides a higher degree of complexity that better approximates real-world experimental setups while still being able to systematically control the variability of the stimuli.

Our interdisciplinary work provides important insights into how multisensory cognitive functions can be modelled in robots operating in crossmodal environments also in conditions of sensory conflict.
We believe that these aspects of multisensory interaction are crucial for the leverage of current models of robot perception which should yield swift and robust behaviour by combining sensory-driven information with internally-generated expectations.

\addtolength{\textheight}{-12cm}   



%

\section*{ACKNOWLEDGMENT}

\small{This research was supported by National Natural Science Foundation of China (NSFC), the China Scholarship Council, and the German Research Foundation (DFG) under project Transregio Crossmodal Learning (TRR 169). 
The authors would like to thank Jonathan Tong, Athanasia Kanellou, Matthias Kerzel, Guochun Yang, and Zhenghan Li for discussions and technical support.}


\balance
\bibliographystyle{IEEEtran}
\bibliography{cmlbiblio18}

\begin{thebibliography}{10}
\providecommand{\url}[1]{#1}
\csname url@samestyle\endcsname
\providecommand{\newblock}{\relax}
\providecommand{\bibinfo}[2]{#2}
\providecommand{\BIBentrySTDinterwordspacing}{\spaceskip=0pt\relax}
\providecommand{\BIBentryALTinterwordstretchfactor}{4}
\providecommand{\BIBentryALTinterwordspacing}{\spaceskip=\fontdimen2\font plus
\BIBentryALTinterwordstretchfactor\fontdimen3\font minus
  \fontdimen4\font\relax}
\providecommand{\BIBforeignlanguage}[2]{{%
\expandafter\ifx\csname l@#1\endcsname\relax
\typeout{** WARNING: IEEEtran.bst: No hyphenation pattern has been}%
\typeout{** loaded for the language `#1'. Using the pattern for}%
\typeout{** the default language instead.}%
\else
\language=\csname l@#1\endcsname
\fi
#2}}
\providecommand{\BIBdecl}{\relax}
\BIBdecl

\bibitem{Noda14}
K.~Noda, H.~Arie, Y.~Suga, and T.~Ogata, ``Multimodal integration learning of
  robot behavior using deep neural networks.'' \emph{Robotics and Autonomous
  Systems}, vol.~62, no.~6, pp. 721 -- 736, 2014.

\bibitem{Vavrecka2013}
M.~Vavre\v{c}ka and I.~Farka\v{s}, ``A multimodal connectionist architecture
  for unsupervised grounding of spatial language,'' \emph{Cognitive
  Computation}, vol.~6, pp. 101--112, 2013.

\bibitem{Hwang17}
J.~Hwang and J.~Tani, ``Seamless integration and coordination of cognitive
  skills in humanoid robots: {A} deep learning approach,''
  \emph{arXiv:1706.02423}, 2017.

\bibitem{Polley2017}
D.~B. Polley, ``Multisensory conflict resolution: {S}hould {I} stay or should
  {I} go?'' \emph{Neuron}, vol.~93, no.~4, pp. 725 -- 727, 2017.

\bibitem{CruzIROS}
F.~Cruz, G.~Parisi, J.~Twiefel, and S.~Wermter, ``Multi-modal integration of
  dynamic audiovisual patterns for an interactive reinforcement learning
  scenario,'' in \emph{Proceedings of the IEEE/RSJ Intl. Conf. on Intelligent
  Robots and Systems (IROS)}, 2016, pp. 759--766.

\bibitem{Malacuso2005}
E.~Macaluso, N.~George, R.~Dolan, C.~Spence, and J.~Driver, ``Spatial and
  temporal factors during processing of audiovisual speech: a pet study,''
  \emph{NeuroImage}, vol.~21, no.~2, pp. 725 -- 732, 2004.

\bibitem{Jack73}
C.~E. Jack and W.~R. Thurlow, ``{Effects of degree of visual association and
  angle of displacement on the "ventriloquism" effect},'' \emph{Perceptual \&
  {M}otor {S}kills}, vol.~37, pp. 967--979, 1973.

\bibitem{mcgurk76}
H.~McGurk and J.~MacDonald, ``Hearing lips and seeing voices,'' \emph{Nature},
  vol. 264, pp. 746--748, 1976.

\bibitem{Kording17}
K.~P. K{\"o}rding, U.~Beierholm, W.~J. Ma, S.~Quartz, J.~B. Tenenbaum, and
  L.~Shams, ``Causal inference in multisensory perception,'' \emph{PLOS ONE},
  vol.~2, no.~9, pp. 1--10, 2007.

\bibitem{Zhu2017}
L.~L. Zhu and M.~S. Beauchamp, ``Mouth and voice: A relationship between visual
  and auditory preference in the human superior temporal sulcus,''
  \emph{Journal of Neuroscience}, 2017.

\bibitem{ParisiICDL}
G.~I. Parisi, P.~Barros, M.~Kerzel, H.~Wu, G.~Yang, Z.~Li, X.~Liu, and
  S.~Wermter, ``A computational model of crossmodal processing for conflict
  resolution,'' in \emph{IEEE International Conference on Development and
  Learning and on Epigenetic Robotics (EPIROB-ICDL)}.\hskip 1em plus 0.5em
  minus 0.4em\relax IEEE, 2017, pp. 33--38.

\bibitem{Dahl2010}
C.~Dahl, N.~Logothetis, and C.~Kayser, ``Modulation of visual responses in the
  superior temporal sulcus by audio-visual congruency,'' \emph{Frontiers in
  Integrative Neuroscience}, vol.~4, p.~10, 2010.

\bibitem{Magosso12}
E.~Magosso, C.~Cuppini, and M.~Ursino, ``A neural network model of
  ventriloquism effect and aftereffect,'' \emph{PLOS ONE}, vol.~7, no.~8, pp.
  1--19, 2012.

\bibitem{Odegaard15}
B.~Odegaard, D.~Wozny, and L.~Shams, ``Biases in visual, auditory, and
  audiovisual perception of space.'' \emph{PLoS Computational Biology},
  vol.~11, no.~12, 2015.

\bibitem{Parise2012}
C.~V. Parise, C.~Spence, and M.~O. Ernst, ``When correlation implies causation
  in multisensory integration,'' \emph{Current Biology}, vol.~22, no.~1, pp. 46
  -- 49, 2012.

\bibitem{Diehl2014}
M.~M. Diehl and L.~M. Romanski, ``Responses of prefrontal multisensory neurons
  to mismatching faces and vocalizations,'' \emph{Journal of Neuroscience},
  vol.~34, no.~34, pp. 11\,233--11\,243, 2014.

\bibitem{Metta2010}
G.~Metta, L.~Natale, F.~Nori, G.~Sandini, D.~Vernon, L.~Fadiga, C.~von Hofsten,
  K.~Rosander, M.~Lopes, J.~Santos-Victor, A.~Bernardino, and L.~Montesano,
  ``The icub humanoid robot: An open-systems platform for research in cognitive
  development,'' \emph{Neural Networks}, vol.~23, no.~8, pp. 1125 -- 1134,
  2010.

\bibitem{Barros2016}
P.~Barros and S.~Wermter, ``Developing crossmodal expression recognition based
  on a deep neural model,'' \emph{Adaptive behavior}, vol.~24, no.~5, pp.
  373--396, 2016.

\bibitem{Luo17}
W.~Luo, Y.~Li, R.~Urtasun, and R.~S. Zemel, ``Understanding the effective
  receptive field in deep convolutional neural networks,'' \emph{CoRR}, vol.
  abs/1701.04128, 2017.

\bibitem{Yang2017}
G.~Yang, W.~Nan, Y.~Zheng, H.~Wu, Q.~Li, and X.~Liu, ``Distinct cognitive
  control mechanisms as revealed by modality-specific conflict adaptation
  effects,'' \emph{Journal of Experimental Psychology: Human Perception and
  Performance}, vol.~43, no.~4, pp. 807--818, 2012.

\end{thebibliography}

\end{document}